\documentclass[conference]{IEEEtran}
\IEEEoverridecommandlockouts

\usepackage{cite}
\usepackage{amsmath,amssymb,amsfonts}
\usepackage{algorithm}
\usepackage{algpseudocode}
\usepackage{graphicx}
\usepackage{textcomp}
\usepackage{xcolor}
\usepackage{booktabs}
\usepackage{listings}
\usepackage{tikz}
\usetikzlibrary{arrows.meta,positioning,backgrounds}

\def\BibTeX{{\rm B\kern-.05em{\sc i\kern-.025em b}\kern-.08em
    T\kern-.1667em\lower.7ex\hbox{E}\kern-.125emX}}
\begin{document}

\title{Bridging the Sim-to-Real Gap in Reinforcement Learning-Based Industrial Dispatching through Execution Semantics
\thanks{Supported by the Chips Joint Undertaking and its members, including top-up funding by National Authorities, within the Cynergy4MIE project (Grant Agreement No. 101140226).}
}

\author{\IEEEauthorblockN{1\textsuperscript{st} Jonathan Hoss}
\IEEEauthorblockA{\textit{Department of Industrial Engineering} \\
\textit{Rosenheim Technical University of Applied Sciences}\\
Rosenheim, Germany \\
jonathan.hoss@th-rosenheim.de}
\and
\IEEEauthorblockN{2\textsuperscript{nd} Noah Klarmann}
\IEEEauthorblockA{\textit{Department of Industrial Engineering} \\
\textit{Rosenheim Technical University of Applied Sciences}\\
Rosenheim, Germany \\
noah.klarmann@th-rosenheim.de}
}

\maketitle

\begin{abstract}
Event-driven scheduling policies are increasingly deployed in industrial environments, where decisions are made under asynchronous and partially observed system states. As a result, decision states are not temporally consistent, action admissibility is not explicitly defined, and the origin of execution errors remains ambiguous. These issues limit both reliability and interpretability.

To address this gap, a policy-neutral execution and measurement layer is proposed to mediate between scheduling policies and the industrial execution environment. The layer constructs decision-valid snapshots from asynchronous event streams, defines a standardized execution contract with explicit action admissibility, and records outcomes as divergences between policy intent, transactional outcomes, physical execution, and human intervention. This enables a separation between decision semantics and execution behavior and makes deployment mismatch observable and structurally attributable.

The proposed framework is evaluated using a discrete-event simulation. The results show analytical benefits across all observation lag regimes, as undifferentiated execution failures are transformed into structured, typed outcomes with full attribution coverage. Operational benefits are strongest under low observation lag, where avoidable execution errors can be prevented before commitment. Overall, the layer turns execution uncertainty into supervisory data for evaluation and policy refinement.
\end{abstract}


\section{Introduction}
The Job-Shop Scheduling Problem (JSSP) serves as the foundational model for formulating and training dispatching policies in production control \cite{ouelhadjSurveyDynamicScheduling2009}. While classical approaches rely on static heuristics such as Priority Dispatching Rules, recent research increasingly adopts Reinforcement Learning (RL) to learn adaptive dispatching behavior directly from simulated JSSP environments \cite{zhangLiteratureReviewReinforcement2025}. Driven by advances in graph-based representations, scalable RL architectures, and emerging offline training methods, recent work suggests that RL-based dispatching can improve policy quality, adaptability, and applicability to larger scheduling instances \cite{smitGraphNeuralNetworks2025, zhangLiteratureReviewReinforcement2025, remmerdenOfflineReinforcementLearning2025,hossScalableProductionScheduling2026}.

Despite this progress, industrial adoption of RL-based dispatching remains limited. As emphasized in recent work on the validation and deployment of RL-based scheduling systems, successful deployment depends not only on optimization performance but also on monitoring, explainability, safety, workflow integration, operator interaction, and sim-to-real robustness \cite{stockermannReinforcementLearningBased2025}. 

In practice, policies are embedded into Manufacturing Execution Systems (MES), Enterprise Resource Planning (ERP) systems, and physical production environments, where they must operate under asynchronous data flows, partial observability, and human intervention. Under these conditions, the scheduler no longer acts on a synchronized and fully observed state. Instead, decisions are based on reconstructed and time-lagged observations derived from heterogeneous event streams. This introduces a structural mismatch between policy assumptions and execution reality, which is not addressed by improvements in policy quality alone \cite{dulacArnoldChallengesRealWorld2019}.

This paper argues that the central limitation lies in the absence of an execution and measurement layer that mediates between decision-making and industrial execution systems. Without such a layer, decision validity, execution feasibility, and outcome attribution remain implicitly defined and cannot be analyzed in a structured way.

To address this limitation, the paper proposes a policy-neutral execution and measurement layer that constructs decision-valid state representations from asynchronous data, defines a shared execution contract across heterogeneous policies, and enables structured attribution of execution outcomes. The contribution is methodological and architectural, focusing on deployment-time semantics rather than policy design.

\section{Sim-to-Real Gap in Current Literature}
Existing work on RL-based dispatching acknowledges the discrepancy between training environments and industrial deployment. The literature can be broadly grouped into three lines of work that address this gap.

One line of research focuses on robustness to disturbances. Approaches in this category train policies under varying system conditions, for instance by incorporating dynamic job arrivals, machine disruptions, stochastic processing times, setup times, transport processes, buffer constraints, or other production-related restrictions \cite{luoDynamicSchedulingFlexible2020, panzerDesigningAdaptiveDeep2024, tesslerActionRobustReinforcement2019, smitGraphNeuralNetworks2025,hossProductionSchedulingFramework2025}. The objective is to improve generalization by exposing the policy to a wider range of scenarios during training.

A second line of work relies on imitation learning and offline learning. These methods learn dispatching behavior directly from historical data or expert demonstrations \cite{remmerdenOfflineReinforcementLearning2025, smitGraphNeuralNetworks2025, dulacArnoldChallengesRealWorld2019}. By leveraging observed execution data, they aim to capture operational patterns that are difficult to model explicitly.

A third line includes meta-learning and knowledge transfer approaches. These methods aim to improve adaptability across environments by transferring knowledge between tasks or compressing policies into more efficient representations \cite{zhaoSimtoRealTransferDeep2020, panzerDesigningAdaptiveDeep2024}. This enables faster adaptation to new production settings without retraining from scratch.

Despite their differences, these approaches share a common strategy. They improve deployment performance by increasing the robustness of the policy with respect to discrepancies between training and execution conditions. At the same time, they largely retain the assumption that a valid decision state and a consistent action interface are available at inference time.

As a result, the underlying causes of deployment mismatch remain mainly unaddressed. The focus lies on absorbing inconsistencies between observation and execution rather than on making them explicit, measurable, and attributable. This limitation motivates the execution and measurement concept proposed in this paper, which shifts the focus from policy-level robustness to explicit execution semantics at deployment time.

\section{Execution-Level Requirements for Industrial Dispatching}
The sim-to-real gap in industrial dispatching reflects a mismatch between scheduling decisions and their realization in production systems. This mismatch arises from structural characteristics of deployment environments that are not explicitly captured in existing approaches.

This paper identifies three execution-level challenges that arise in industrial deployment. These concern state construction under asynchronous observations, execution semantics and action admissibility across heterogeneous systems, and the structured attribution of deviations between intended and realized execution. In the following, the requirements for the design of the execution and measurement layer are derived from these challenges.

\subsection{State Construction under Asynchronous Observations}
The deployment of dispatching policies in industrial environments is fundamentally constrained by the way the state of the system is constructed at decision time \cite{dulacArnoldChallengesRealWorld2019}. In contrast to the synchronized and fully observable state representations typically assumed in JSSP formulations and RL training environments, real-world production systems operate on asynchronous, event-driven data streams. Information about job progress, machine status, and resource availability is distributed across MES and ERP systems and often arrives with non-negligible delays or inconsistencies. As a result, the scheduler-facing state is not a direct representation of the underlying real-world system but a reconstructed state assembled from partial and temporally misaligned observations \cite{xuSurveyIndustrialInternet2018}.

This mismatch shifts the effective problem formulation from a fully observable Markov decision process to a partially observable setting. As a result, the scheduler operates on a state that does not necessarily correspond to the underlying physical and transactional system state. Decisions that are valid with respect to the available observation may therefore be infeasible at execution time due to delayed updates, missing confirmations, or unobserved system changes.

This projection gap requires that the execution and measurement layer proposed in this paper provide an explicit mechanism for constructing decision-valid state representations, integrating asynchronous inputs, enforcing temporal consistency, and representing uncertainty arising from incomplete or delayed information.

\subsection{Action Admissibility under Execution Constraints}
In addition to the projection of incomplete and temporally misaligned system states, a further challenge in industrial deployment arises from the lack of an explicit definition of how decisions are interpreted and executed. Current scheduling approaches implicitly rely on a well-defined action space and a set of admissible decisions that are available at inference time. In industrial settings, this assumption is not guaranteed \cite{dulacArnoldChallengesRealWorld2019}. While frameworks such as ISA-95 define hierarchical data exchange between enterprise and control levels \cite{martinezAutomationPyramidConstructor2021}, and MES and ERP systems manage transactional workflows and system coordination, neither provides a scheduler-facing definition of when a decision is valid, how it is triggered, or under which conditions an action can be executed \cite{xuSurveyIndustrialInternet2018}.

Even in emerging Industry 4.0 architectures, where infrastructures such as IoT and digital twins improve system visibility and representation \cite{xuSurveyIndustrialInternet2018,lvDigitalTwinsIndustry2023}, the focus remains on data availability rather than on how decisions are translated into system actions \cite{qiuEdgeComputingIndustrial2020}. 

This gap becomes particularly relevant when integrating heterogeneous scheduling approaches within the same production environment, as these policies rely on different implicit assumptions about state completeness, timing, and action feasibility while operating on the same underlying system. Without a shared definition of how decisions are interpreted, candidate actions remain ambiguous and cannot be consistently validated prior to execution.

This leads to the second requirement for the execution and measurement layer proposed in this paper. The layer must define an execution contract that specifies how decisions are represented, when they are considered valid, how they are triggered, and how they are translated into system actions. It must provide explicit semantics for action admissibility and ensure consistent validation, rejection, or deferral of decisions under transactional constraints. In addition, it must enable the integration of human interventions as explicitly represented elements of the execution process.

\subsection{Structuring Execution Outcome Attribution}
A further challenge in industrial deployment concerns the attribution of execution outcomes. In current practice, deviations between planned and executed actions are not attributed to distinct causes. Failed or altered actions are recorded as generic disturbances, without systematically distinguishing whether deviations arise from physical constraints, transactional inconsistencies, or human intervention. At the same time, the transition to Industry 5.0 increasingly foregrounds human-in-the-loop decision making \cite{lvDigitalTwinsIndustry2023, destouetFlexibleJobShop2023}. As a result, human interventions should no longer be treated as exogenous noise, but must be explicitly represented as a distinct source of execution outcomes.

The lack of differentiation limits evaluation and learning. Observed outcomes do not reliably reflect policy quality, as deviations may result from external constraints rather than suboptimal decisions. Historical data is likewise unsuitable for learning, since different types of execution failure are conflated.

This leads to the requirement that the execution and measurement layer must provide a structured representation linking policy intent, execution context, and realized behavior. It must explicitly distinguish sources of deviation, thereby enabling consistent evaluation and systematic policy improvement.

\section{Proposed Architecture: The Execution and Measurement Layer}

We propose an execution and measurement layer that acts as a middleware between dispatch policies and the industrial execution environment. It controls how heterogeneous physical and transactional event streams are translated into decision-relevant states and defines when and under which conditions a policy is invoked. The following sections specify the mechanisms of state isolation via snapshots, decision formulation through execution contracts, and outcome attribution, as well as the implementation constraints required for deployment.

\subsection{State Snapshot Isolation for Decision Validity}
One requirement derived from the sim-to-real problem is the need to construct a decision-valid system state at inference time in industrial environments with heterogeneous and asynchronous data sources.
Continuously updating the policy state would expose the agent to fractured and temporally inconsistent information. The proposed layer addresses this by introducing a snapshot isolation mechanism that provides a consistent view of the system and defines how decision-relevant state is constructed and exposed under asynchronous conditions.

Incoming physical and transactional event streams are accumulated in a continuously updated execution cache. When a dispatch trigger occurs, the layer isolates a point-in-time snapshot $s_k$ that represents the scheduler-facing system state at decision epoch $k$. This isolation resolves conflicting timestamps and integrates both physical observations and transactional annotations required for execution.

Once isolated, the snapshot remains immutable throughout the decision process. The policy evaluates $s_k$ without exposure to concurrent state updates. This prevents inconsistent state transitions during inference and ensures that the decision is based on a temporally stable representation. However, isolation does not imply completeness. Events that have not yet arrived or transactional states that remain latent in surrounding systems are not represented in $s_k$, and may affect the validity of the decision at execution time.

To maintain consistency between decision and execution, the layer monitors incoming events for critical updates that invalidate the current decision context. If such an event occurs before the intended action is committed, the decision is aborted, the snapshot is discarded, and a new isolation step is initiated. Repeated invalidation can lead to a livelock condition under sustained event volatility. In practice, this is mitigated by bounding inference latency, restricting the set of events that trigger invalidation, and introducing fallback dispatch modes such as heuristic or operator-controlled decisions.

With this snapshot isolation mechanism, the policy is invoked on a chronologically valid state that reflects all information available at the time of decision while explicitly handling the limitations of asynchronous observability.

\subsection{Policy-Neutral Execution Contract}
To address the requirement of ensuring action admissibility under execution constraints, the layer defines a policy-neutral, event-driven execution contract that specifies how decisions are formulated and which actions are admissible at inference time. Rather than exposing an implicit or system-dependent action space, it explicitly defines the decision interface through a structured request.

At each decision epoch $k$, triggered when a resource becomes dispatchable-idle, the layer formulates the decision request
\[
R_k = \langle s_k, m_k, J_k^{cand} \rangle,
\]
where $s_k$ denotes the isolated state snapshot, $m_k$ the idle resource, and $J_k^{cand}$ the candidate set defined by the layer. The candidate set is constructed as
\[
J_k^{cand} = J_k^{phys} \cap J_k^{sys},
\]
that is, the intersection of physically feasible and transactionally admissible actions under the snapshot $s_k$.

This construction separates physical feasibility from transactional admissibility. Physical observations may indicate that an action is executable on the shop floor, but such observations can be incomplete or temporally misaligned. Conversely, MES and ERP systems encode the transactional conditions under which an action is formally admissible, including release state, routing consistency, reservation status, and related constraints. By defining admissible actions as the intersection of these two sets, the layer ensures that only actions that are both physically plausible and transactionally valid are exposed to the policy.

The snapshot $s_k$ is a scheduler-facing observation rather than complete ground truth. In live operation, the candidate set represented in $s_k$ may differ from the true executable set in the plant due to asynchronous event arrival. The policy evaluates $R_k$ and selects an intended action $a_k^{intent} \in J_k^{cand}$ based on this representation. This action may subsequently be rejected or found infeasible during execution when a previously latent state becomes observable. The layer translates $a_k^{intent}$ into a standards-compliant transactional payload and submits it to the execution system while recording the intended action separately from the resulting execution outcome.

By standardizing both the structure of decision requests and the semantics of admissible actions, the execution contract ensures that policy outputs are expressed on a consistent basis across different dispatch strategies. Because decisions are formulated against a common candidate definition and snapshot semantics, the contract provides a stable and comparable interface for evaluating policy behavior at deployment time.

\subsection{Divergence Representation and Outcome Attribution}
Building on the standardized execution contract, the proposed layer defines how execution outcomes are recorded to enable structured attribution at deployment time. Rather than collapsing execution into a single observed outcome, the layer introduces a typed representation that separates policy intent from system-level and physical execution results.

For each decision epoch $k$, the layer records the divergence tuple
\[
D_k = \langle a_k^{intent}, x_k^{sys}, x_k^{phys}, u_k \rangle,
\]
where $a_k^{intent}$ denotes the action selected by the policy, $x_k^{sys}$ the transactional system response, $x_k^{phys}$ the physical execution outcome, and $u_k$ a potential human intervention. Each component corresponds to a distinct execution channel with its own semantics.

This representation separates intended and realized execution across system boundaries. The transactional outcome $x_k^{sys}$ captures whether the action was accepted, rejected, or modified by MES or ERP systems, while $x_k^{phys}$ reflects the actual shop-floor outcome and $u_k$ records manual overrides or corrections. Together, these channels form a divergence tuple that prevents execution outcomes from being collapsed into a single failure signal.

This divergence tuple makes deviations attributable to distinct causes. In particular, it distinguishes between visible and hidden divergence. In the visible case, the blocking condition is already represented in the snapshot $s_k$, so an inadmissible or suboptimal action is attributable to the policy. In the hidden case, the blocking condition exists in the execution environment but is absent from $s_k$ due to delayed or missing information. This separates policy error from limitations in observability and system synchronization and provides a reusable supervisory record for downstream refinement, including improved admissibility modeling, offline learning, and reuse of human interventions.

Algorithm~\ref{alg:execution-layer} summarizes the end-to-end execution and measurement process, including event ingestion, decision execution, and divergence recording.

\begin{algorithm}[t]
\caption{Dispatch through the execution layer}
\label{alg:execution-layer}
\begin{algorithmic}[1]
\State Receive multi-source event $e_t$
\State Update execution cache with $e_t$
\If{$e_t$ opens a valid dispatch trigger}
    \State Latch decision snapshot $s_k$
    \State Compute admissible candidate set $J_k^{cand}$
    \State Invoke policy with $R_k = \langle s_k, m_k, J_k^{cand} \rangle$
    \State Record policy intent $a_k^{intent}$
    \State Await transactional response and record $x_k^{sys}$
    \State Await physical outcome and record $x_k^{phys}$
    \If{human override occurs}
        \State Record intervention $u_k$
    \EndIf
    \State Emit divergence record $D_k$
\EndIf
\end{algorithmic}
\end{algorithm}

\subsection{Implementation Constraints}
A practical deployment cannot assume that all decision-relevant information is available from a single synchronized source. Instead, the proposed layer must integrate heterogeneous execution and transactional systems under bounded latency conditions. On the execution side, the layer ingests machine and sensor events from physical resources. On the transactional side, it incorporates MES and ERP updates, such as job release status, routing validity, reservations, material availability, or quality holds. These inputs must be normalized into a unified event stream before being processed by the snapshot isolation mechanism.

A central requirement is the separation of physical event ingestion, transactional state updates, and scheduler-facing decision exposure. This separation ensures that asynchronous updates from different system layers do not directly interfere with the construction of decision-valid snapshots. Instead, all incoming information is first incorporated into a continuously updated execution cache, from which isolated decision states are derived.

The snapshot isolation mechanism further imposes constraints on system latency. To maintain decision validity, the delay between event arrival, state update, snapshot formation, and policy invocation must remain bounded relative to the dispatch horizon of the application. This does not require hard real-time guarantees, but instead implies that inference operates on a sufficiently fresh state to remain operationally relevant.

From a systems perspective, the continuously updated execution state is maintained in a low-latency in-memory store. In contrast, divergence tuples $D_k$ and their associated event histories are persisted as analytical artifacts outside the control path. This separation ensures that decision latency is not affected by logging or storage operations while still enabling reproducibility and offline analysis.

\section{Empirical Evaluation of Execution Semantics}
To evaluate the proposed execution semantics, we choose a controlled simulation setting that isolates execution-layer effects. The objective is not to benchmark scheduling performance but to expose the impact of the execution layer independently of the underlying dispatch policy. The following presents the simulation setup and experimental design, followed by an evaluation of decision validity, the impact of execution contracts on action admissibility, the structured attribution of execution outcomes, and a summary across observation lag regimes.

\subsection{Simulation Setup and Experimental Design}
The study is conducted using a discrete-event simulation implemented in SimPy \cite{simpy}. The simulation uses a single non-preemptive machine as the dispatchable resource, with dynamic job arrivals triggering decision epochs whenever the machine becomes idle. It enforces a strict separation between ground truth and scheduler-visible state, where, for each waiting job, transactional and physical readiness are represented both as true state variables and as delayed observable variables.

The single-machine setting is deliberately chosen as a minimal evaluation environment. Its purpose is not to reproduce the full complexity of industrial scheduling, but to isolate the effect of execution semantics under controlled asynchronous observability. Multi-machine routing, resource coupling, and precedence constraints would introduce additional scheduling effects that could obscure whether improvements originate from the execution layer or from the dispatching logic. The evaluation is therefore designed as a proof of concept for the execution and measurement mechanism rather than as a complete validation of industrial scheduling performance.

\begin{table}[t]
\centering
\caption{Simulation configuration for the empirical evaluation}
\label{tab:simulation-config}
\small
\setlength{\tabcolsep}{4pt}
\begin{tabular}{@{}ll@{}}
\toprule
\textbf{Component} & \textbf{Configuration} \\
\midrule
Processing times & $\mathrm{Uniform}(3,8)$ \\
Interarrival times & $\mathrm{Uniform}(5.5,10.5)$ \\
Disturbance probabilities & $p_{sys}=0.14$, $p_{phys}=0.10$, $p_{hum}=0.07$ \\
Decision window & $0.85$ time units \\
Lag presets & Low: $\mathrm{Uniform}(0.0,0.3)$ \\
& Medium: $\mathrm{Uniform}(0.1,1.5)$ \\
& High: $\mathrm{Uniform}(0.5,3.0)$ \\
Replications & 50 per condition \\
Simulation horizon & 2000 time units \\
Warm-up fraction & 0.05 \\
\bottomrule
\end{tabular}
\end{table}

Jobs are characterized by arrival time, processing time, due date, weight, transactional readiness, and physical readiness. The model separates ground-truth plant state from \mbox{scheduler-visible} state. Execution disturbances are injected through temporary transactional blocks, temporary physical faults, and human overrides, parameterized by $p_{sys}$, $p_{phys}$, and $p_{hum}$. The corresponding simulation parameters are summarized in Table~\ref{tab:simulation-config}.

Scenarios are constructed by varying the observation lag to capture different levels of asynchronous observability. Observation lag is defined as the delay between a change in the underlying system state and its visibility to the scheduler. This delay is interpreted relative to the decision window, defined as the time between state observation and action commitment. In the simulation, this decision window spans $0.85$ time units. Observation lags shorter than this window allow updates to become visible before a decision is committed, whereas longer delays cause updates to remain hidden until after execution.

Based on this relation, three lag regimes are considered. Their numerical definitions are given in Table~\ref{tab:simulation-config}. The low-lag setting models cases in which most updates become visible before commitment. The medium-lag setting straddles the decision window and represents partial observability at commit time. The high-lag setting models delayed propagation, where relevant updates frequently arrive only after commitment, increasing the share of hidden execution failures.

To isolate architectural effects from learning dynamics, two standard heuristics, Earliest Due Date (EDD) and Shortest Processing Time (SPT), are evaluated under both a direct-execution baseline and the proposed execution layer. Both execution approaches dispatch from the observable set of admissible candidates. However, only the execution layer invalidates pending decisions when critical updates arrive before commit and preserves typed execution outcomes.

\subsection{Evaluation of Decision Validity}

\begin{table}[htbp]
\centering
\caption{Performance comparison under medium observation lag}
\label{tab:sim-support}
\small
\setlength{\tabcolsep}{4pt} 
\begin{tabular}{@{}lrr@{}}
\toprule
\textbf{Metric} & \textbf{Direct Exec.} & \textbf{Exec. Layer} \\ 
\midrule
Invalid dispatches\textsuperscript{a} & $61.81 \pm 1.31$ & $48.68 \pm 1.19$ \\
Visible divergence\textsuperscript{a}  & $15.90 \pm 0.62$  & $0.00$            \\
Mean weighted tardiness\textsuperscript{a}     & $11.21 \pm 1.55$ & $6.43 \pm 0.88$ \\
Throughput\textsuperscript{a}         & $0.1268 \pm 0.0003$ & $0.1267 \pm 0.0003$ \\
\bottomrule
\multicolumn{3}{@{}p{\columnwidth}@{}}{\footnotesize \textsuperscript{a} Means with 95\% CI ($50$ replications per policy, $n=100$ after aggregation over EDD and SPT) at medium lag.
}
\end{tabular}
\end{table}

Table~\ref{tab:sim-support} summarizes the simulation results for the medium-lag scenario, reporting mean values per run. Relative to direct execution, the execution layer reduces invalid dispatches from 61.81 to 48.68. This indicates that decisions are more frequently based on a valid system state when snapshot isolation and pre-commit invalidation are applied.

\begin{figure}[t]
    \centering
    \includegraphics[width=\columnwidth]{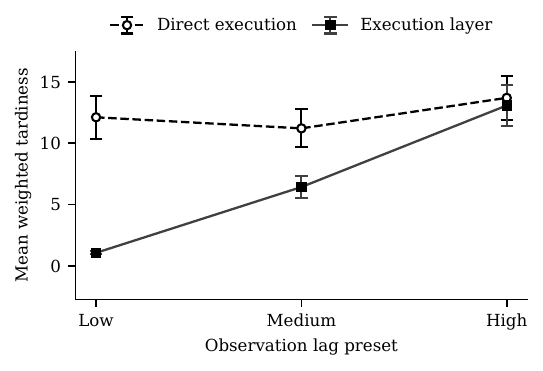} 
    \caption{Mean weighted tardiness under increasing observation lag for direct execution and the execution layer.}
    \label{fig:robustness-plot}
\end{figure}

The effect of decision validity under varying levels of asynchronous observability is further illustrated in Fig.~\ref{fig:robustness-plot}. To quantify the operational impact of invalid decisions, mean weighted tardiness is used as the primary metric, defined as
\[
T_w = \frac{1}{N} \sum_{i=1}^{N} w_i \max(0, C_i - d_i),
\]
where $C_i$, $d_i$, and $w_i$ denote the completion time, due date, and priority weight of job $i$, respectively. Under direct execution, mean weighted tardiness remains relatively stable across lag regimes (12.11, 11.21, 13.69), indicating that stale-information effects persist across the evaluated observation settings. In contrast, the execution layer yields substantially lower tardiness at low lag (1.08) and still lower tardiness at medium lag (6.43). At high lag, the difference becomes small (13.07 versus 13.69), indicating that the operational benefit diminishes once relevant updates no longer arrive within the actionable decision window.

This behavior confirms that decision validity is primarily determined by whether relevant state changes become observable within the decision window. When this condition is met, the layer prevents avoidable errors and offers an operational benefit. When it is not, execution converges toward the behavior of direct execution.

\subsection{Impact of Execution Contracts on Action Admissibility}
The execution contract is designed to enforce explicit action admissibility, thereby preventing visible divergence by construction. As shown in Table~\ref{tab:sim-support} for the medium-lag scenario, visible divergence amounts to $15.90 \pm 0.62$ under direct execution, indicating that actions are committed despite already observable conflicts. In contrast, the execution layer reduces visible divergence to 0.00 by invalidating pending decisions when critical updates arrive before commitment.

At the same time, no measurable throughput penalty is observed for the execution layer. Throughput remains effectively unchanged across architectures and lag settings. This is consistent with the intended role of the execution contract, which does not alter the underlying scheduling objective or system capacity but only constrains when and under which conditions actions are committed.

These metrics show that the execution contract enforces admissibility without introducing policy-dependent bias or affecting system throughput.

\subsection{Structured Attribution of Execution Outcomes}

Direct execution collapses outcomes into undifferentiated failure signals, resulting in no attribution coverage. Attribution coverage measures the fraction of outcomes that can be assigned to a distinct causal category. In contrast, the execution layer, through its explicit divergence representation, enables the assignment of outcomes to distinct categories such as transactional blocking, physical constraints, or human intervention, thereby achieving full attribution coverage.

\begin{figure}[t]
    \centering
    \includegraphics[width=0.85\columnwidth]{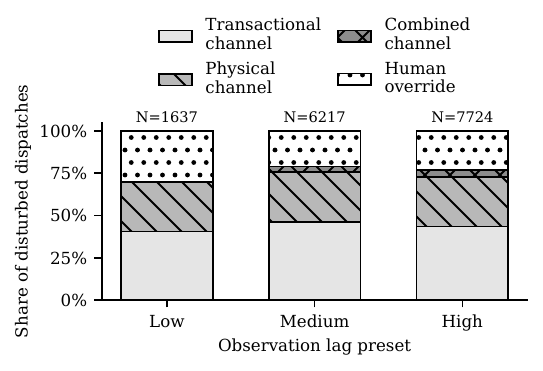}
    \caption{Learner-visible composition of disturbed dispatches by outcome type under varying observation lag for the execution layer. Here, $N$ denotes the number of disturbed dispatches represented in each bar.}
    \label{fig:divergence-composition}
\end{figure}

Figure~\ref{fig:divergence-composition} illustrates the learner-visible composition of disturbed dispatches by outcome type under varying observation lag for the execution layer. The figure shows that transactional, physical, combined, and human-override categories are preserved across all regimes, maintaining a structured representation of execution behavior.

This structured representation provides a richer and more informative supervisory signal for downstream refinement, including disturbance-aware value estimation, improved admissibility modeling, and reuse of human interventions, because it allows the underlying causes to be distinguished. While the execution layer establishes the basis for this increased informational content, the granularity of the resulting signal depends on the chosen level of outcome categorization. In practice, this introduces a trade-off between operational feasibility and the desired level of causal resolution, as more fine-grained distinctions can provide additional insight but may require more detailed instrumentation and integration.

\subsection{Summary Across Lag Regimes}
Across all lag regimes, the execution layer exhibits a consistent pattern. At low observation lag, where updates become visible within the decision window, the layer provides a clear operational advantage by preventing invalid decisions and reducing tardiness. At medium lag, this benefit is reduced but remains observable. At high lag, where updates arrive predominantly after commitment, operational behavior converges toward direct execution, as invalidation can no longer prevent incorrect actions.

However, the analytical advantages of the execution layer persist across all regimes. The layer maintains full attribution coverage and structured outcome representation. This ensures that execution behavior remains interpretable, transforming deployment mismatch into explicit and reusable execution evidence.

Although the empirical evaluation uses a single-machine setting, the proposed execution semantics are not limited to this case. The layer generalizes to multi-machine environments by applying the same execution semantics per dispatchable resource, while extending $J_k^{cand}$ to routing constraints, machine eligibility, setup dependencies, buffer availability, reservation conflicts, and synchronization requirements. The divergence tuple $D_k$ remains structurally unchanged, but may require additional attribution categories for routing infeasibility, inter-machine blocking, missing predecessor completion, buffer saturation, or synchronization conflicts. Thus, the main scalability challenge lies in constructing fresh and consistent snapshots across coupled resources, rather than in the policy interface itself.

\section{Conclusion and Future Work}
We introduce a policy-neutral execution and measurement layer as an architectural prerequisite for deployable event-driven industrial dispatch. The contribution of this paper is twofold. First, we identify three structural challenges underlying the sim-to-real gap at deployment time: the lack of decision-valid state representations under asynchronous observation, the absence of explicit action admissibility, and the confounded attribution of execution outcomes. Second, we propose a policy-neutral execution and measurement layer that enforces stable snapshot isolation during decision making, defines a standardized execution contract with explicit action admissibility, invalidates decisions upon critical updates before commitment, and records a typed separation between policy intent and realized execution outcomes. Rather than introducing another integration interface, the layer establishes execution semantics that make deployment mismatch observable, reusable, and comparable across heterogeneous dispatch policies.

The empirical evaluation demonstrates two complementary effects. Operationally, the layer reduces invalid dispatches and improves tardiness most clearly when delayed updates still arrive within the actionable decision window, while throughput remains essentially unchanged. As observation lag increases, this operational advantage diminishes, as a larger fraction of disturbances remains hidden until commit time. Analytically, however, the layer transforms otherwise undifferentiated execution failures into structured, typed outcomes and thereby enables explicit attribution of execution behavior. Because decisions are evaluated under a consistent snapshot semantics and execution contract, this representation further supports a comparable interpretation of execution outcomes across different dispatch policies. This attribution remains valuable even under severe lag, where the primary benefit shifts from operational correction toward improved interpretability and diagnosability of execution outcomes.

Future work will extend the framework beyond simulation by integrating the layer into a physical execution environment with MES connectivity, focusing on latency and robustness under realistic conditions. A central next step is the integration of the divergence tuple $D_k$ into post-deployment learning pipelines to enable offline reinforcement learning, disturbance-aware value estimation, and the systematic reuse of human interventions for model improvement.

\end{document}